\documentclass{article}
\usepackage{spconf,amsmath,graphicx,xcolor}
\usepackage{subfigure}
\usepackage{cite}
\usepackage{amsmath}
\usepackage{algorithmic}
\usepackage{array}
\usepackage{url}

\usepackage{amssymb}
\usepackage{multirow}
\usepackage{tabu}
\usepackage{pbox}

{\begin{list}               
    {$\bullet$ \hfill}{
        \setlength{\leftmargin}{\parindent}
        \setlength{\parsep}{0.04\baselineskip}
        \setlength{\itemsep}{0.5\parsep}
        \setlength{\labelwidth}{\leftmargin}
        \setlength{\labelsep}{0em}}
    }
{\end{list}}

\providecommand{\cref}[1]{Chapter~\ref{#1}}

\providecommand{\R}{\ensuremath{\mathbb{R}}}

\providecommand{\abs}[1]{\lvert#1\rvert}
\providecommand{\norm}[1]{\lVert#1\rVert}

\renewcommand{\vec}[1]{\ensuremath{\boldsymbol{#1}}}
\providecommand{\mat}[1]{\ensuremath{\boldsymbol{#1}}}

\providecommand{\calD}{\mathcal{D}}

\providecommand{\calT}{\mathcal{T}}

\providecommand{\mD}{\mat{D}}

\providecommand{\vu}{\vec{u}}
\providecommand{\vv}{\vec{v}}

\providecommand{\vx}{\vec{x}}

\providecommand{\ptilde}{\widetilde{p}}

\newcommand{\argmin}[1]{\mathop{\underset{#1}{\mbox{argmin}}}}

\newcommand{\etal}{{\it{et al}.} }

\title{Hand Segmentation for Hand-Object Interaction from Depth Map}

\name{Byeongkeun~Kang$^{\star}$ Kar-Han~Tan$^{\dagger}$ Nan~Jiang$^{\star}$ Hung-Shuo~Tai$^{\dagger}$ Daniel~Tretter$^{\ddagger}$ Truong~Nguyen$^{\star}$\thanks{This work is supported in part by NSF grant IIS-1522125.}}

\address{$^{\star}$ Department of Electrical and Computer Engineering, UC San Diego, La Jolla, CA 92093 USA \\
    $^{\dagger}$ NovuMind Inc., Santa Clara, CA 95054 USA \\
    $^{\ddagger}$ Hewlett-Packard, Inc., Palo Alto, CA 94304 USA}

\begin{document}

\maketitle

\begin{abstract}
Hand segmentation for hand-object interaction is a necessary preprocessing step in many applications such as augmented reality, medical application, and human-robot interaction. However, typical methods are based on color information which is not robust to objects with skin color, skin pigment difference, and light condition variations. Thus, we propose hand segmentation method for hand-object interaction using only a depth map. It is challenging because of the small depth difference between a hand and objects during an interaction. To overcome this challenge, we propose the two-stage random decision forest (RDF) method consisting of detecting hands and segmenting hands. To validate the proposed method, we demonstrate results on the publicly available dataset of hand segmentation for hand-object interaction. The proposed method achieves high accuracy in short processing time comparing to the other state-of-the-art methods. 
\end{abstract}

\begin{keywords}
Hand segmentation, human-machine interaction, random decision forest, depth map
\end{keywords}

\section{Introduction}
Recently, with the expansion of virtual reality (VR), augmented reality (AR), robotics, and intelligent vehicles, the development of new interaction technologies has become unavoidable since these applications require more natural interaction methods rather than input devices. For these applications, many researches have been conducted such as gesture recognition and hand pose estimation. However, most technologies focus on understanding interactions which do not involve touching or handling any real world objects although understanding interactions with objects is important in many applications. We believe that this is because hand segmentation is much more difficult in hand-object interaction. Thus, we present a framework of hand segmentation for hand-object interaction. 

\subsection{Related work} \label{subsec:relatedWork}
Hand segmentation has been studied for many applications such as hand pose estimation~\cite{tompson, sharp, qian,romeroicra,romeroivc, wangtog}, hand tracking~\cite{oikoiccv, argyros,kangisvc}, and gesture/sign/grasp recognition~\cite{cai,kangacpr}. In color image-based methods, skin color-based method has been popular~\cite{jones, khan, li, cai, phung, kakumanu}. For hand-object interaction, Oikonomidis \etal and Romero \etal segmented hands by thresholding skin color in HSV space~\cite{oikoiccv, romeroicra, romeroivc, argyros}. Wang \etal processed hand segmentation using a learned probabilistic model where the model is constructed from the color histogram of the first frame~\cite{wangtog}. Tzionas \etal applied skin color-based segmentation using the Gaussian mixture model~\cite{tzionas}. However, skin color-based segmentation has limitations in interacting with objects in skin color, segmenting from other body parts, skin pigment difference, and light condition variations. An alternative method is wearing a specific color glove~\cite{wang}. 

For depth map-based methods, popular methods are using a wrist band~\cite{kangacpr, kangisvc, qian} or using random decision forest (RDF)~\cite{tompson, shottoncvpr, sharp}. Although the method using a black wristband is uncomplicated and effective, it is inconvenient. Moreover, the method cannot segment hands from objects during hand-object interaction since it processes segmentation by finding connected components. Tompson \etal\cite{tompson} and Sharp \etal\cite{sharp} proposed the RDF-based methods based on~\cite{shottoncvpr}. Although the purposes of the methods are slightly different comparing to the proposed method, the methods are the most relevant methods. 

In this paper, we propose the hand segmentation method for hand-object interaction using only a depth map to avoid the limitations of skin color-based methods. We present the two-stage RDF method to achieve high accuracy efficiently. 

\begin{figure}[!t]
\centering
   \includegraphics[width=0.45\textwidth]{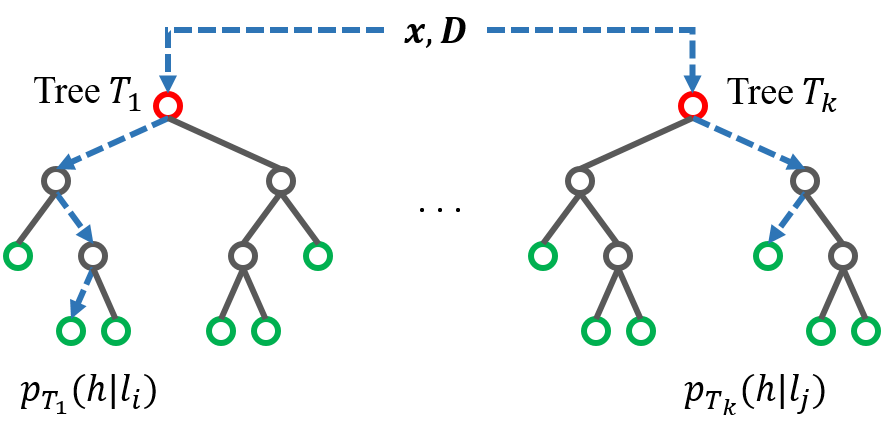}
   \caption{Random decision forest. Red, black, and green circles represent root nodes, split nodes, and leaf nodes, respectively. }
\label{fig:RDF}
\end{figure}

\section{Method} \label{sec:method}
We propose two-stage RDF for hand segmentation for hand-object interaction. In our two-stage RDF, the first RDF detects hands by processing the RDF on an entire depth map. Then, the second RDF segments hands in pixel-level by applying the RDF in the detected region. This cascaded architecture is designed for the second RDF to focus on the segmentation of hands from objects and close body parts such as an arm.

RDF consists of a collection of decision trees as shown in Fig.~\ref{fig:RDF}. Each decision tree is composed of a root node, splitting nodes, and leaf nodes. Given an input data at the root node, it is classified to child nodes based on the split function at each splitting node until it reaches a leaf node. In this paper, the input data is the location of each pixel on a depth map. The split function uses the feature of the depth difference between two relative points on the depth map in~\cite{shottoncvpr}. At a leaf node, a conditional probability distribution is learned in a training stage, and the learned probability is used in a testing stage. For more details about RDF, we refer the readers to~\cite{criminisi, Breiman2001, ho1995}.

\subsection{Training} \label{subsec:RDFtraining}
\begin{figure}[!t]
\centering
    \subfigure{\includegraphics[width=0.23\textwidth]{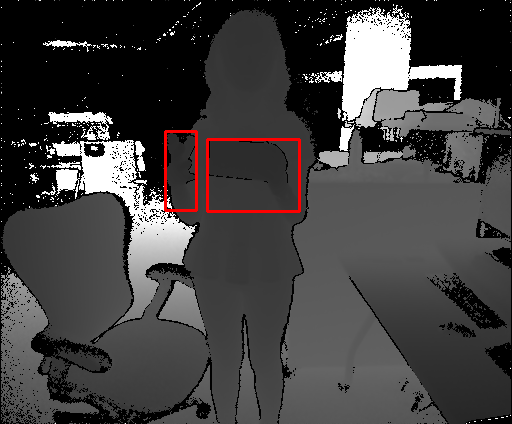}}
    \subfigure{\includegraphics[width=0.23\textwidth]{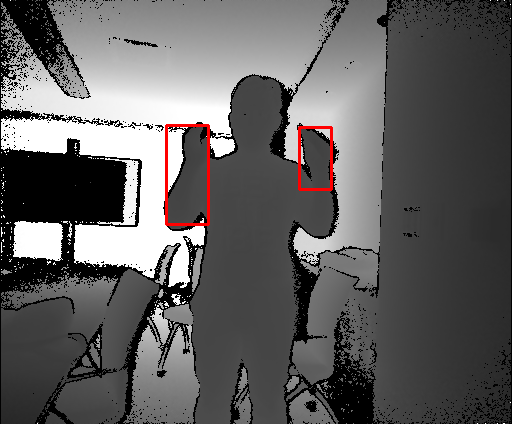}}
   \caption{Detection of hands using the RDF in the first stage.}
\label{fig:boundingbox}
\end{figure}

Given a training dataset $\calD$, the algorithm randomly selects a set $\calD_i$ of depth maps $\mD$ and then randomly samples a set of data points $\vx$ in the region of interest (ROI) on the selected depth maps $\mD$. The ROI is the entire region of the depth maps in the first stage. It is the detected regions using the first RDF in the second stage (see Fig.~\ref{fig:boundingbox}). The sampled set of data points $\vx$ and the corresponding depth maps $\mD$ are inputs to the training of a decision tree.

Using the inputs ($\vx$, $\mD$), the algorithm learns a split function at each splitting node and a conditional probability distribution at each leaf node. First, learning the split function includes learning a feature $f(\cdot)$ and a criteria $\theta$. We use the feature $f(\cdot)$ of the depth difference between two relative points $\{ \vx+ {\vu}/{\mD_{\vx}}, \vx+{\vv}/{\mD_{\vx}} \}$ in~\cite{shottoncvpr} as follows:
\begin{equation}
f(\vx, \mD, \vu, \vv) = D_{\vx+{\vu}/{\mD_{\vx}} } - D_{\vx+{\vv}/{\mD_{\vx}} }
\label{eq:feature}
\end{equation}
where $\mD_{\vx}$ denotes the depth at a pixel $\vx$ on a depth map $\mD$; $\vu \in \R^{2}$ and $\vv \in \R^{2}$ represent offset vectors for each relative point. Then, the criteria $\theta$ decides to split the data $\vx$ to the left child or the right child.
\begin{equation}
f(\vx, \mD, \vu, \vv)  \lessgtr \theta.
\end{equation}
Thus, the algorithm learns two offset vectors ($\vu, \vv$) and a criteria $\theta$ at each splitting node. 

Since the goal is separating the data points $\vx$ of different classes to different child nodes, the objective function is designed to evalutate the separation using the learned offset vectors and criteria as follows:
\begin{equation}
L(\vx, \mD, \vu, \vv, \theta) = -\sum_{c\in\{l,r\}}\sum_{h\in\{0,1\}}\frac{|\vx_c|}{|\vx|} p(h|c) \log{p(h|c)}
\label{eq:split}
\end{equation}
where $c$ and $h$ are indexes for child nodes $\{l, r\}$ and for classes, respectively; $|\vx_c|$ denotes the number of data points in the $c$ child node; $p(h|c)$ is the estimated probability of being the class $h$ at the child node $c$.

To learn offsets and a criteria, the algorithm randomly generates possible candidates and selects the candidate with a minimum loss $L(\cdot)$ as follows: 
\begin{equation}
(\vu, \vv, \theta) = \argmin {(\vu, \vv, \theta)} {L(\vx, \mD, \vu, \vv, \theta)}.
\end{equation}

Learning a split function at each splitting node is repeated until the node satisfies the condition for a leaf node. The condition is based on (1) the maximum depth of the tree, (2) the probability distribution $p(h|c)$, and (3) the amount of training data $|\vx|$ at the node. Specifically, it avoids too many splitting nodes by limiting the maximum depth of the tree and by terminating if the child node has a high probability for a class or if the amount of remaining training data is too small. 

At each leaf node, the algorithm stores the conditional probability $p(h|l)$ (probability of being each class $h$ given reaching the node $l$) for the prediction in a testing stage.

\subsection{Testing} \label{subsec:decision}
Using the learned RDF, the algorithm predicts the probability of being a class for a new data $\vx$. The new data is classified to child nodes using the learned split function at each splitting node until it reaches a leaf node. At the leaf node $l$, the learned conditional probability $p_{T}(h|l)$ is loaded. These steps are repeated for entire trees $T$ in the forest $\calT$. Then, the probabilities are averaged to predict the probability $p(h|\vx)$ of being a class $h$ for the new data $\vx$.
\begin{equation}
p(h|\vx) = \frac{1}{|\calT|} \sum_{T \in \calT} {p_{T}(h|l)}
\end{equation}
where $|\calT|$ is the number of trees in the learned forest $\calT$.  

In the first stage, the first RDF is applied on an entire depth map to compute a probability map. Then, the probability map is used to detect hands as shown in Fig.~\ref{fig:boundingbox}. In the second stage, the second RDF processes the data points in the detected regions to predict the probability of being each class. The proposed two-stage RDF improves both accuracy and efficiency by focusing on each task in each stage.

\begin{figure}[!t]
\centering
\subfigure[]{\includegraphics[width=0.239\textwidth]{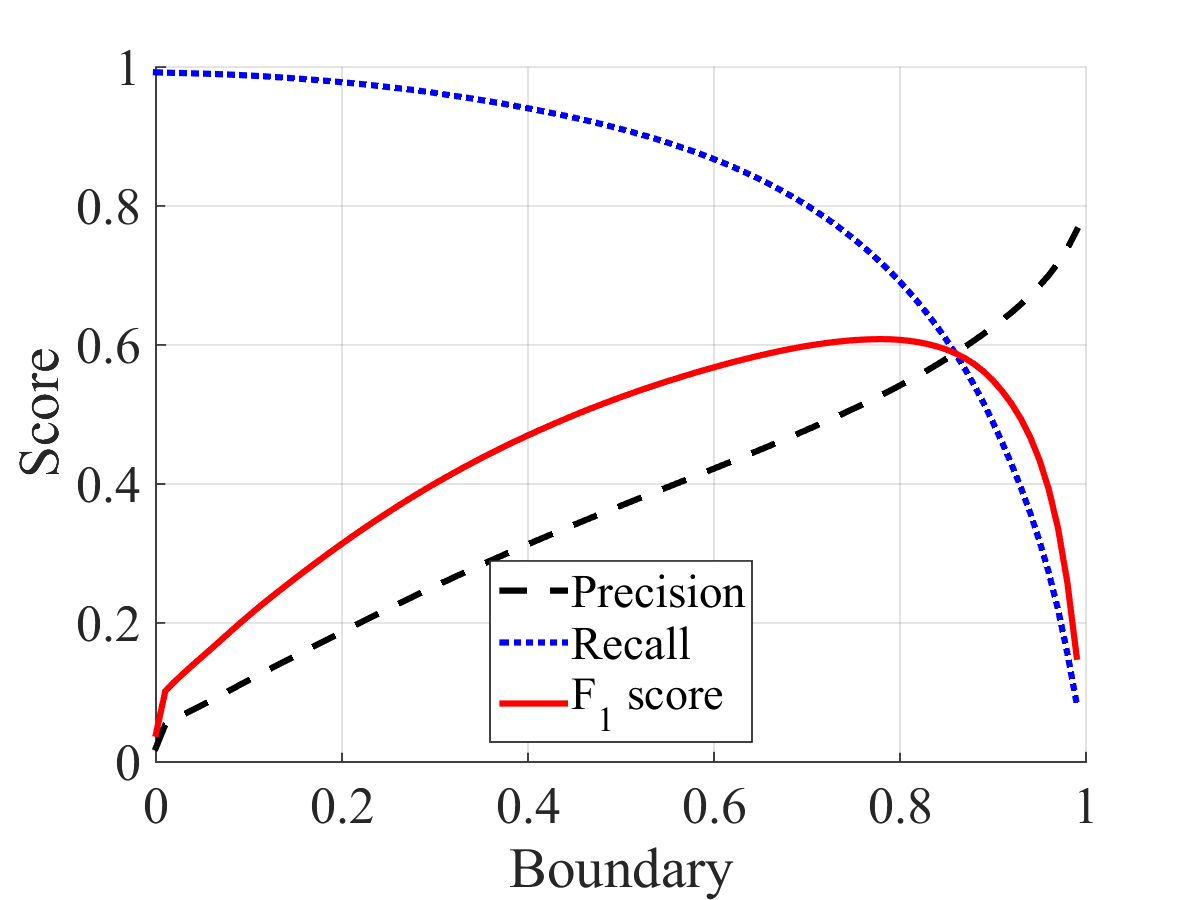}}
\subfigure[]{\includegraphics[width=0.239\textwidth]{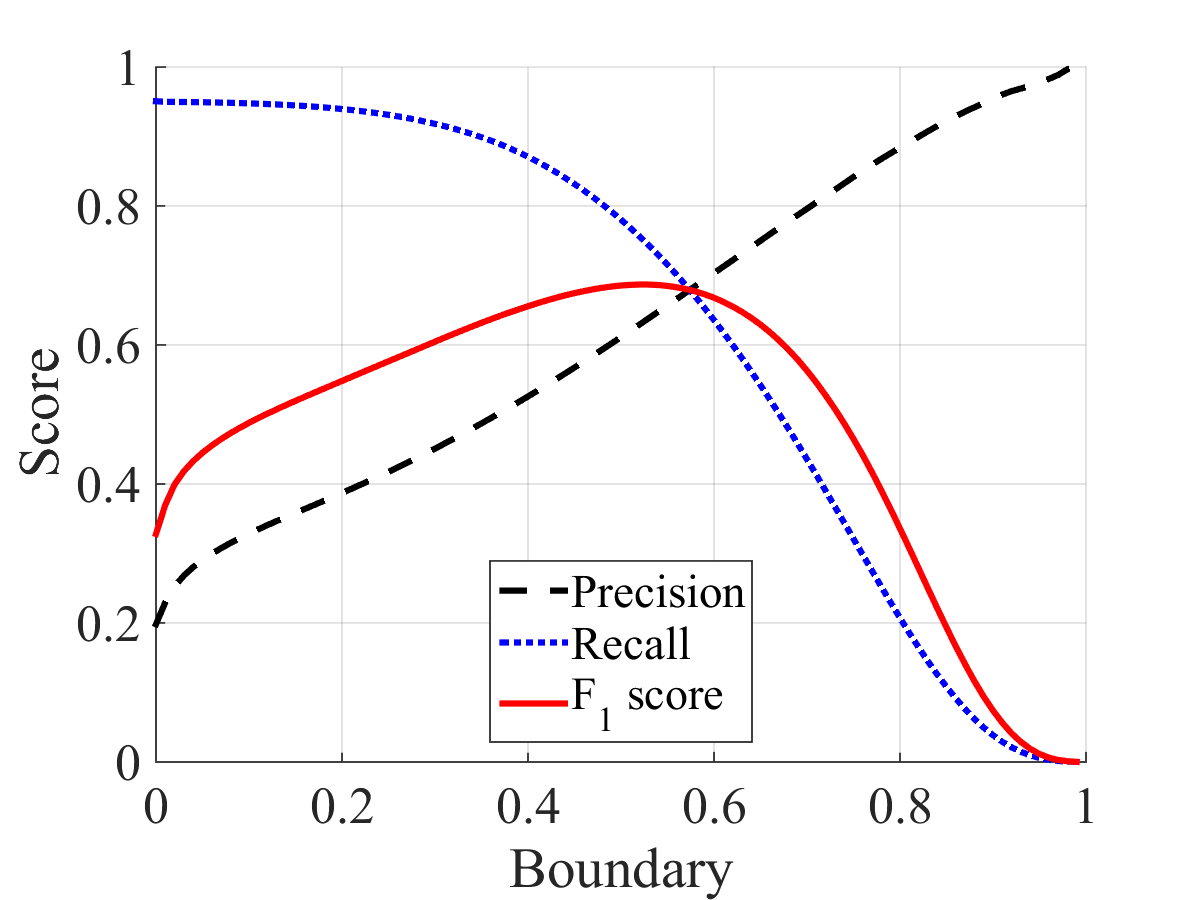}}
   \caption{Scores depending on the decision boundary on the validation dataset. (a) Score of the RDF in the first stage. (b) Score of the two-stage RDF with filtering in Section~\ref{subsec:filtering}.}
\label{fig:decision}
\end{figure}

Decision boundaries are exhaustively searched with the step size of 0.01 using the predicted probability maps of the validation dataset as shown in Fig.~\ref{fig:decision}. Although the most typical boundary is 0.5 for a probability map, we found that it is not the best parameter. The selected boundaries are shown in Table~\ref{tab:result}.

\subsection{Modified bilateral filter} \label{subsec:filtering}
Before classifying a data $\vx$ to a class $h$, modified bilateral filter is applied to the predicted probability $p(h|\vx)$ to make the probability more robust. Since the probability $p(h|\vx)$ is predicted for each pixel independently, the probability is stabilized by averaging the probabilities of the data points in close distance and similar intensity on the depth map.


Unlike typical bilateral filter whose weights are based on the input image (in this case, the probability map)~\cite{tomasi}, the weights in the modified bilateral filter are based on a separate image, the depth map. The filtering is defined as follows:
\begin{equation}
\ptilde(h|\vx) = \frac{1}{w} \sum_{\vx_i \in \Omega}  g_r( \abs{\mD_{\vx_i} - \mD_{\vx}}) g_s(\norm{\vx_i - \vx})  p(h|\vx_i)
\end{equation}
where $\Omega$ is the set of pixels within the filter's radius and the pre-defined depth difference; $w$ is the normalization term, 
$w = \sum_{\vx_i \in \Omega}  g_r( \abs{\mD_{\vx_i} - \mD_{\vx}}) g_s(\norm{\vx_i - \vx})$;
$g_r(\cdot)$ and $g_s(\cdot)$ are the Gaussian functions for the depth difference and for the spatial distance from the data point $\vx$, respectively.
$g_r(r) = \exp (-\frac{r^2}{2\sigma_r^2}); g_s(s) = \exp (-\frac{s^2}{2\sigma_s^2} ).$
The parameters in the filter were selected based on the experiments using validation dataset. 
The selected parameters are as follows: the maximum depth difference to be considered is $400mm$. Both standard deviations ($\sigma_r$ and $\sigma_s$) are 100.


\begin{figure}[!t]
\centering
    \includegraphics[width=0.45\textwidth]{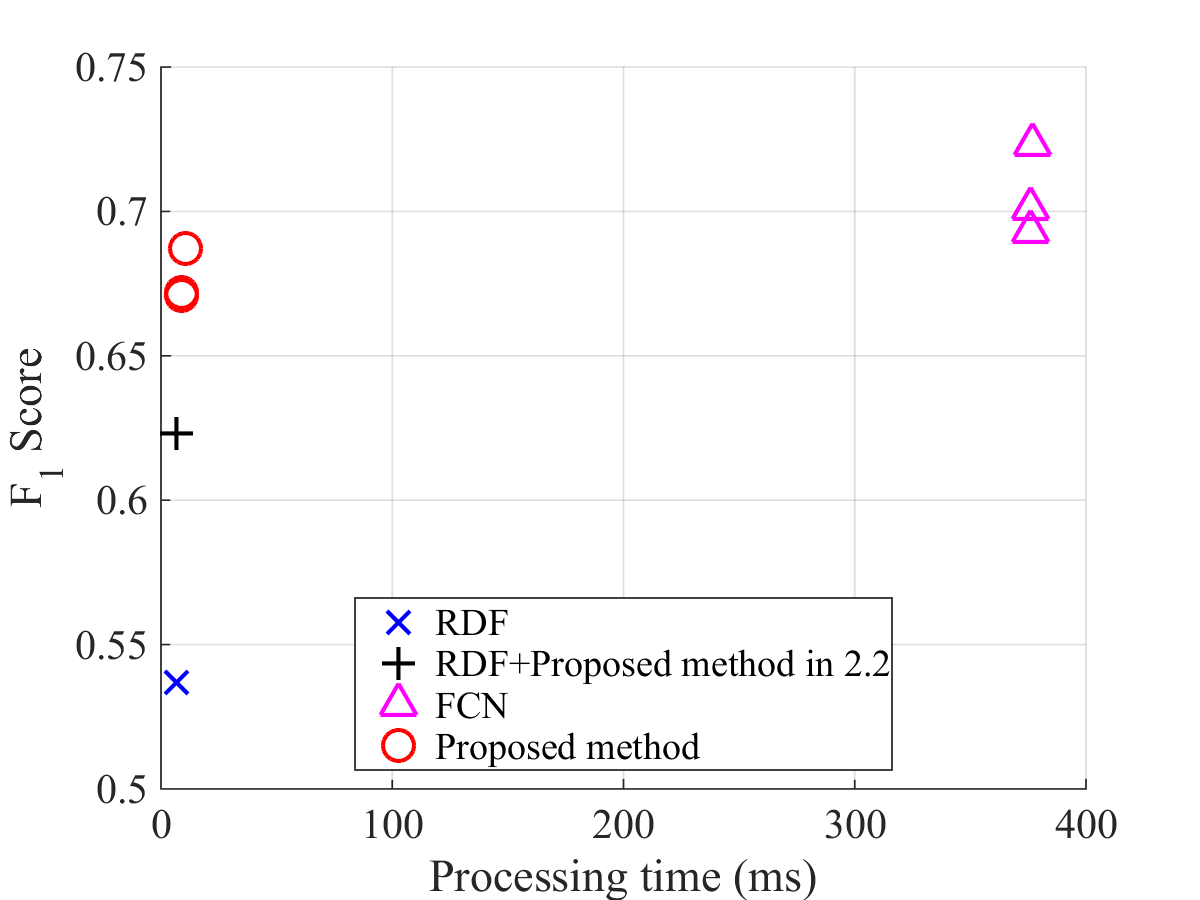}
   \caption{Analysis of accuracy and efficiency.}
\label{fig:accuracyEfficiency}
\end{figure}


\begin{table*}[!t]
\caption{Quantitative comparison. The two boundaries for the proposed method are for each stage.}
\label{tab:result}
\centering
\renewcommand{\arraystretch}{1.2}
\begin{tabu} to 0.96\textwidth {c|c|X[c,m]|X[c,m]|X[c,m]|X[c,m]|c }
\hline
\multicolumn{3}{c|}{Method} & \multicolumn{3}{c|}{Score} & Processing time \\
\cline{1-6}
Method & Boundary & Filter & Precision & Recall & $F_1$ score & ($ms$) \\
\hline\hline
{RDF~\cite{tompson, shottoncvpr}} 				& 0.50 & - 		& 38.1 & 91.2 & 53.7 & 6.7  \\
\hline
{RDF~\cite{tompson, shottoncvpr}} + Proposed in Sec.~\ref{subsec:decision}	& 0.78	& -  & 54.5 & 72.7 & 62.3 & 6.7  \\
\hline
{FCN-32s~\cite{longCVPR, longPAMI}} 				& - 	& - 	& 70.0 & 68.6 & 69.3 & 376  \\
{FCN-16s~\cite{longCVPR, longPAMI}} 				& - 	& - 	& 68.0 & 72.2 & 70.1 & 376  \\
{FCN-8s~\cite{longCVPR, longPAMI}} 				& - 	& - 	& 70.4 & 74.4 & 72.3 & 377  \\
\hline
\multirow{3}{*}{Proposed method}					& 0.50, 0.50	& - & 59.2 & 77.4 & 67.1 & 8.9  \\
\cline{2-7}
 	     													& 0.50, 0.52	& - & 60.8 & 75.1 & 67.2 & 8.9 \\
\cline{2-7}
 	     													& 0.50, 0.52	& 11 $\times$ 11 & 62.9 & 75.6 & 68.7 & 10.7 \\
\hline
\end{tabu}
\end{table*}

\section{Experimental Evaluations} \label{sec:result}
\subsection{Dataset} \label{subsec:dataset}
We collected a new dataset\footnote{\url{https://github.com/byeongkeun-kang/HOI-dataset}} using Microsoft Kinect v2~\cite{KangTMM}. The newly collected dataset consists of 27,525 pairs of depth maps and ground truth labels from 6 people (3 males and 3 females) interacting with 21 different objects. Also, the dataset includes the cases of one hand and both hands in a scene. The dataset is separated into 19,470 pairs for training, 2,706 pairs for validation, and 5,349 pairs for testing, respectively. 


 \begin{figure*}[!t] \begin{center}
\begin{minipage}{0.19\linewidth}
\centerline{\includegraphics[scale=0.34]{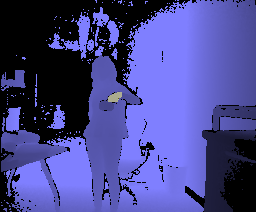}}
\vspace{0.1cm}
\end{minipage}
\begin{minipage}{0.19\linewidth}
\centerline{\includegraphics[scale=0.34]{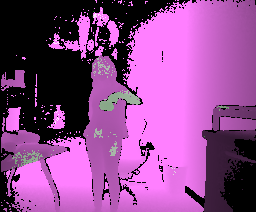}}
\vspace{0.1cm}
\end{minipage}
\begin{minipage}{0.19\linewidth}
\centerline{\includegraphics[scale=0.34]{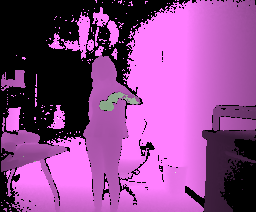}}
\vspace{0.1cm}
\end{minipage}
\begin{minipage}{0.19\linewidth}
\centerline{\includegraphics[scale=0.34]{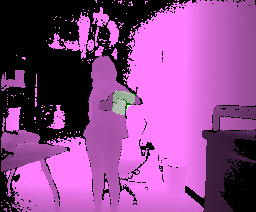}}
\vspace{0.1cm}
\end{minipage}
\begin{minipage}{0.19\linewidth}
\centerline{\includegraphics[scale=0.34]{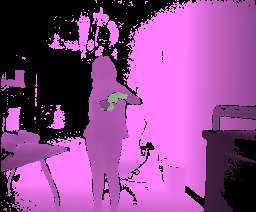}}
\vspace{0.1cm}
\end{minipage}
\\
\begin{minipage}{0.19\linewidth}
\centerline{\includegraphics[scale=0.34]{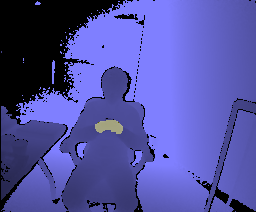}}
\vspace{0.1cm}
\end{minipage}
\begin{minipage}{0.19\linewidth}
\centerline{\includegraphics[scale=0.34]{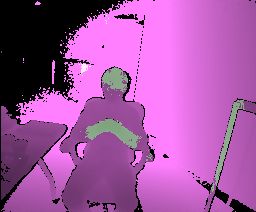}}
\vspace{0.1cm}
\end{minipage}
\begin{minipage}{0.19\linewidth}
\centerline{\includegraphics[scale=0.34]{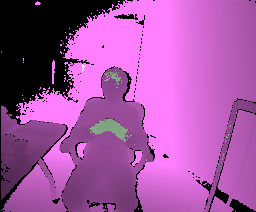}}
\vspace{0.1cm}
\end{minipage}
\begin{minipage}{0.19\linewidth}
\centerline{\includegraphics[scale=0.34]{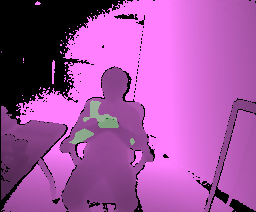}}
\vspace{0.1cm}
\end{minipage}
\begin{minipage}{0.19\linewidth}
\centerline{\includegraphics[scale=0.34]{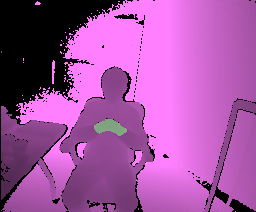}}
\vspace{0.1cm}
\end{minipage}
\\
\begin{minipage}{0.19\linewidth}
\centerline{\includegraphics[scale=0.34]{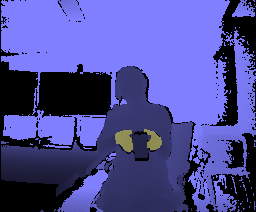}}
\centerline{\footnotesize (a)} 
\end{minipage}
\begin{minipage}{0.19\linewidth}
\centerline{\includegraphics[scale=0.34]{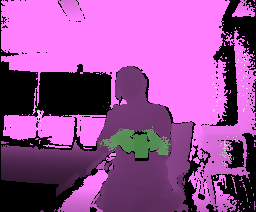}}
\centerline{\footnotesize (b)} 
\end{minipage}
\begin{minipage}{0.19\linewidth}
\centerline{\includegraphics[scale=0.34]{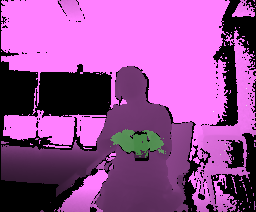}}
\centerline{\footnotesize (c)} 
\end{minipage}
\begin{minipage}{0.19\linewidth}
\centerline{\includegraphics[scale=0.34]{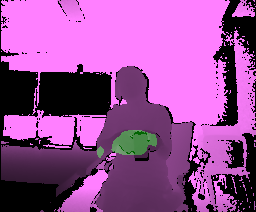}}
\centerline{\footnotesize (d)} 
\end{minipage}
\begin{minipage}{0.19\linewidth}
\centerline{\includegraphics[scale=0.34]{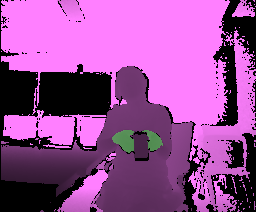}}
\centerline{\footnotesize (e)} 
\end{minipage}
   \caption{Visual comparison. (a) Ground truth label. (b) Result using RDF~\cite{tompson, shottoncvpr}. (c) Result using RDF~\cite{tompson, shottoncvpr} with the proposed method in Section~\ref{subsec:decision}. (d) Result using FCN-8s~\cite{longCVPR, longPAMI}. (e) Result using the proposed method. The results and ground truth label are visualized using different color channels for better visualization.}
\label{fig:result}
\end{center}\end{figure*}

\subsection{Results}
The proposed method is analyzed by demonstrating the results on the dataset in Section~\ref{subsec:dataset}. For the quantitative comparison of accuracy, we measure $F_1$ score, precision, and recall as follows:
\begin{equation}
\begin{split}
\text{precision} &= \frac{\text{tp}}{\text{tp} + \text{fp}}, \quad    
\text{recall} = \frac{\text{tp}}{\text{tp} + \text{fn}}        \\
F_1 &= 2 \times \frac{\text{precision} \times \text{recall}}{\text{precision} + \text{recall}}
\end{split}
\end{equation}
where tp, fp, and fn represent true positive, false positive, and false negative, respectively.
For the comparison of efficiency, we measure the processing time using a machine with Intel i7-4790K CPU and Nvidia GeForce GTX 770. 

The proposed method is compared with the RDF-based method in~\cite{tompson, shottoncvpr} and the fully convolutional networks (FCN) in~\cite{longCVPR,longPAMI} using only a depth map. The proposed method is not compared with color-based methods since the characteristics of depth sensors and color imaging sensors are quite different. For example, a captured depth map using a depth sensor does not vary depending on light condition. However, a captured color image varies a lot depending on light condition. Thus, choosing the capturing environment affects the comparison of results using depth maps and color images. Hence, we only compare the proposed method with the state-of-the-art methods which can process using only depth maps.

Table~\ref{tab:result} and Fig.~\ref{fig:result} show quantitative results and visual results. The scores in Table~\ref{tab:result} are scaled by a factor of 100. The quantitative results show that the proposed method achieves about 25\% and 8\% relative improvements in $F_1$ score comparing to the RDF-based methods~\cite{tompson, shottoncvpr} and its combination with the proposed method in Section~\ref{subsec:decision}, respectively. Comparing to the deep learning-based methods~\cite{longCVPR,longPAMI}, the proposed method achieves about 7\% lower accuracy, but processes in about 42 times shorter processing time. Thus, deep learning-based methods can not be used in real-time applications. Fig.~\ref{fig:accuracyEfficiency} shows the comparison of methods in accuracy and efficiency. The proposed method achieves high accuracy in short processing time.

\section{Conclusion} \label{sec:conclusion}
In this paper, we present two-stage RDF method for hand segmentation for hand-object interaction using only a depth map. The two stages consist of detecting the region of interest and segmenting hands. The proposed method achieves high accuracy in short processing time comparing to the state-of-the-art methods.  

\bibliographystyle{IEEEbib}
\bibliography{refs}

\end{document}